\documentclass[sigconf, table]{acmart}

\usepackage{amsmath,amsfonts}
\usepackage{algorithmic}
\usepackage{algorithm}
\usepackage{array}
\usepackage[caption=false,font=normalsize,labelfont=sf,textfont=sf]{subfig}
\usepackage{textcomp}
\usepackage{stfloats}
\usepackage{url}
\usepackage{verbatim}
\usepackage{acmart-taps}
\usepackage{graphicx}
\usepackage{pifont}
\usepackage{booktabs}
\usepackage{multirow}
\usepackage{makecell}

\usepackage{xcolor,colortbl}

\definecolor{gray}{RGB}{0.89,0.89,0.89}

\usepackage{array}
\definecolor{Gray}{gray}{0.9}
\definecolor{myblue1}{RGB}{144, 201, 230}
\definecolor{myblue2}{RGB}{155, 187, 225}
\definecolor{myblue3}{RGB}{122, 187, 219}
\definecolor{myblue}{RGB}{146, 220, 247}
\definecolor{mycyan}{RGB}{151, 251, 241}
\definecolor{mygreen}{RGB}{220, 254, 141}
\definecolor{mygreen2}{RGB}{237, 254, 198}
\definecolor{myblue4}{RGB}{196, 225, 255}
\definecolor{myblue5}{RGB}{36,80,138}
\definecolor{under}{RGB}{116,141,164}
\usepackage{booktabs}
\usepackage{hyperref}
\AtBeginDocument{%
  }

\copyrightyear{2026}
\acmYear{2026}
\setcopyright{cc}
\setcctype{by}
\acmConference[ICMR '26]{International Conference on Multimedia Retrieval}{June 16--19, 2026}{Amsterdam, Netherlands}
\acmBooktitle{International Conference on Multimedia Retrieval (ICMR '26), June 16--19, 2026, Amsterdam, Netherlands}
\acmDOI{10.1145/3805622.3810684}
\acmISBN{979-8-4007-2617-0/2026/06}

\begin{document}

\title{Toward General and Robust LLM-enhanced Text-attributed Graph Learning}

\author{Zihao Zhang}
\affiliation{%
  \institution{Beijing Institute of Technology}
  \city{Beijing}
  \country{China}
}
\email{zihaozhang648@gmail.com}

\author{Xunkai Li}
\authornote{Corresponding author.}
\affiliation{%
  \institution{Beijing Institute of Technology}
  \city{Beijing}
  \country{China}
}
\email{cs.xunkai.li@gmail.com}

\author{Rong-Hua Li}
\affiliation{%
 \institution{Beijing Institute of Technology}
  \city{Beijing}
  \country{China}
}
\email{lironghuabit@126.com}

\author{Zhenjun Li}
\affiliation{%
  \institution{Shenzhen Institute of Technology}
  \city{Shenzhen}
  \country{China}
}
\email{lizhenjun@szsit.edu.cn}

\author{Bing Zhou}
\affiliation{%
  \institution{Shenzhen City Polytechnic}
  \city{Shenzhen}
  \country{China}
}
\email{zhoubing@szcp.edu.cn}

\author{Guoren Wang}
\affiliation{%
  \institution{Beijing Institute of Technology}
  \city{Beijing}
  \country{China}
}
\email{wanggrbit@126.com}

\begin{abstract}
Recent advancements in Large Language Models (LLMs) and the proliferation of Text-Attributed Graphs (TAGs) across various domains have positioned LLM-enhanced TAG learning as a critical research area. However, the field faces significant challenges: (1) the absence of a unified framework to systematize the diverse optimization perspectives, and (2) the lack of a robust method capable of handling real-world TAGs, which often suffer from text and edge sparsity, leading to suboptimal performance.
To address these challenges, we propose UltraTAG, a unified pipeline for LLM-enhanced TAG learning. UltraTAG provides a comprehensive and domain-adaptive framework that not only organizes existing methodologies but also paves the way for future advancements. Building on this framework, we propose UltraTAG-S, a robust instantiation designed to tackle sparsity issues in real-world TAGs. UltraTAG-S employs LLM-based text propagation and augmentation to mitigate text sparsity, while leveraging LLM-augmented node selection based on PageRank and edge reconfiguration strategies to address edge sparsity. Our experiments demonstrate UltraTAG-S significantly outperforms existing baselines, achieving improvements of 2.12\% and 17.47\% in ideal and sparse settings, respectively. Moreover, as the data sparsity ratio increases, the performance improvement of UltraTAG-S also rises. 
\end{abstract}


\maketitle
\vspace{-4pt}
\section{Introduction}
In recent years, the rapid advancements in large language models (LLMs)~\cite{GPT-3} have significantly driven the continuous evolution of graph ML, particularly in the area of Text-Attributed Graphs (TAGs)\cite{TAPE}, which effectively combine nodes, edges, and textual data for a wide range of applications in domains such as social networks, recommendation systems, and knowledge graphs. While graph neural networks (GNNs)\cite{li2024lightdic} excel at capturing complex structural information, they frequently struggle with textual data, thereby necessitating the integration of GNNs and LLMs for more comprehensive TAG learning\cite{ENGINE,SimTeG}. Despite notable progress, existing TAG learning methods still face several critical limitations:

\textbf{\textit{Limitation 1: Lack of a unified LLM-enhanced TAG learning framework.}}
As the current innovation directions of LLM-enhanced TAG learning remain relatively disorganized and fragmented, we recapitulate them here from a new perspective:
(1) {\textit{Preprocessing: Data Augmentation}}~\cite{TAPE,LLMGNN,GraphBridge,CIKM'24_LLM_as_Teacher}, which leverages LLMs to generate enhanced textual representations like soft labels for text augmentation.
(2) {\textit{Feature Engineering: Improved Text Encoder}}\cite{GIANT,SimTeG}, which utilizes LLMs/LMs to provide more expressive and semantically informative node representations.
(3) {\textit{Training: Joint Training Mechanism}}~\cite{GLEM,ENGINE,G2P2,GraphAdapter}, which further improves performance by establishing an interactive and mutually beneficial training mechanism between GNNs and LMs.
However, the diverse optimization strategies and goals without a systematic standard hinder unified objectives, slowing progress in TAG learning.

\begin{figure*}[!htb]
\centering
\includegraphics[width=1.0\linewidth]{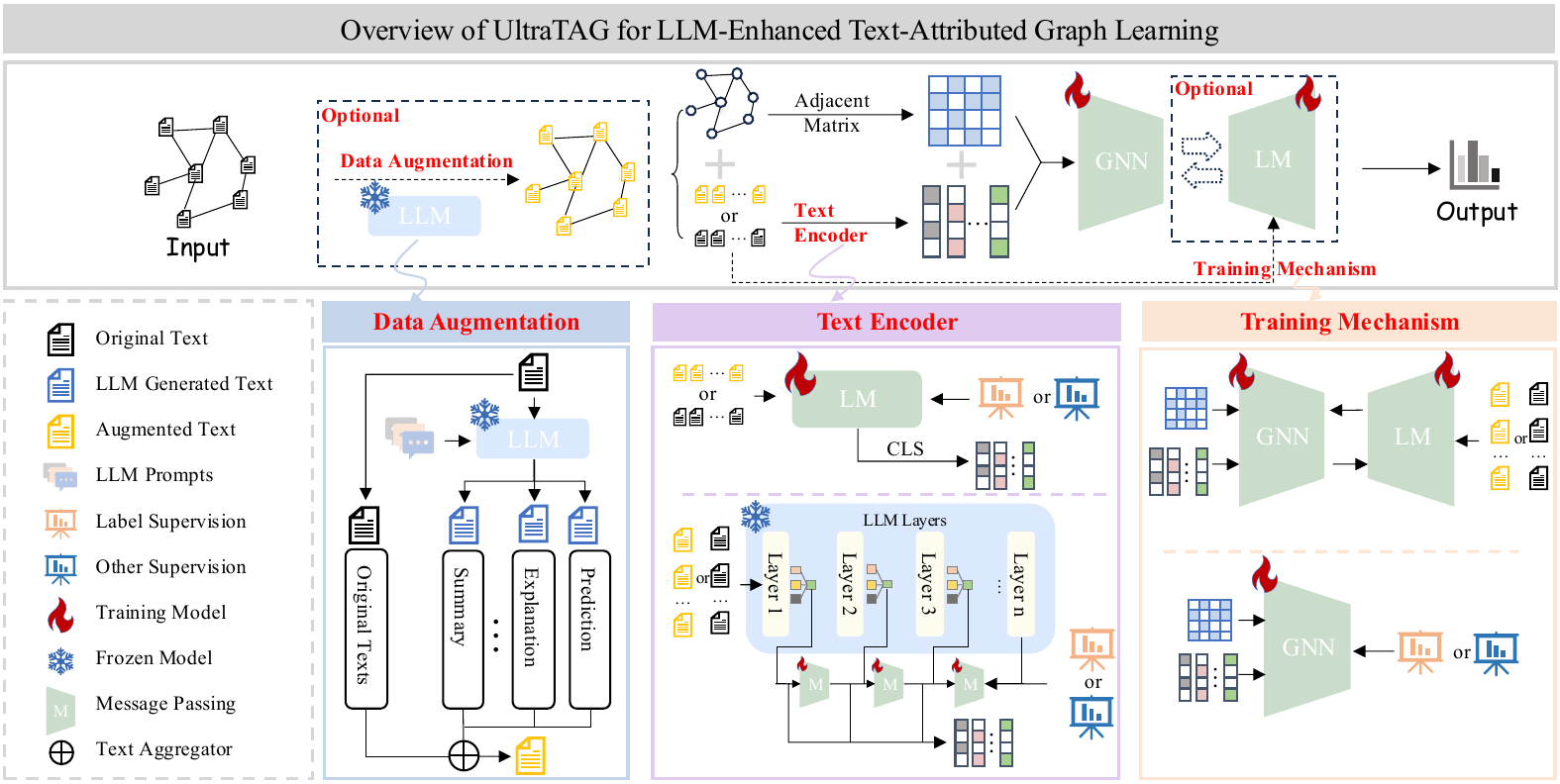}
\caption{Overview of UltraTAG for LLM-Enhanced Text-Attributed Graph Learning with three modules.}
\label{fig:model1}
\end{figure*}

\textbf{\textit{Solution 1: UltraTAG: A \underline{U}nified Pipe\underline{l}ine \underline{t}owa\underline{r}d Gener\underline{a}l and Robust LLM-enhanced \underline{TAG} Learning.}} 
To address Limitation 1, we propose UltraTAG, as detailed in Sec.~\ref{sec:current method}. UltraTAG is composed of three modules: Data Augmentation, Text Encoder, and Training Mechanism, as shown in Figure~\ref{fig:model1}. 
These modules collectively encompass the three directions of LLM-enhanced TAG learning, integrating them cohesively. The different innovative directions are effectively translated into optimization objectives corresponding to the respective modules of UltraTAG. More importantly, UltraTAG is highly adaptable to various real-world scenarios. Building on this flexibility, we introduce UltraTAG-X, a versatile extension for specific challenges. The ``X'' represents a tailored adaptation to unique scenarios, such as data sparsity (UltraTAG-S), high noise levels (UltraTAG-N), or dynamic graph structures (UltraTAG-D). Through its unifying design and extensibility, UltraTAG not only addresses current limitations but also provides a foundation for future innovations, setting a new standard for TAG learning.

\textbf{\textit{Limitation 2: Lack of a Robust Method.}} 
In real-world TAGs, data sparsity in both nodes and edges is a common challenge. For instance, in social networks, privacy measures~\cite{li2024privacypreservinggraphembeddingbased} may limit access to users' information or obscure their connections. Designing a robust method that handles such scenarios with minimal performance degradation presents a significant challenge. The robustness considered here is specific to sparse data, characterized by the absence of a certain proportion of node text attributes and edges. However, most existing approaches~\cite{TAPE,GLEM} rely heavily on complete text attributes for all nodes. When some nodes' text attributes are missing, the model's input structure is disrupted, rendering sparse graphs incompatible with the model's requirements~\cite{SimTeG,LLMGNN}, eventually leading to catastrophic performance decline.

\textbf{\textit{Solution 2: UltraTAG-S: An Instance of UltraTAG for Sparse Scenarios.}} To address Limitation 2, we propose UltraTAG-S, as detailed in Sec.~\ref{sec:our method}. UltraTAG-S is composed of three key modules: (1)LLM-based Robustness Enhancement, (2)LM-based Resilient Representation Learning, and (3)Graph-Enhanced Robust Classifier, as illustrated in Figure~\ref{fig:model2}. To simulate real-world sparse scenarios, we randomly remove node texts and edges from the graph according to a certain ratio. Module 1 includes edge-based text propagation and LLM-based text augmentation to address node sparsity, PageRank-based node selector and LLM-based edge reconfigurator to handle edge sparsity. Module 2 uses LM to get enhanced node embedding. Module 3 incorporates a joint graph structure learning module to further enhance robustness.

\textbf{\textit{Our Contributions:}} 
(1) \underline{\textit{A Unified Framework.}} 
We adopt a novel perspective to systematically examine all existing methods for TAG learning and introduce UltraTAG, a unified and domain-adaptive paradigm which can extend to UltraTAG-X.
(2) \underline{\textit{A Robust Method.}} 
Expanding on UltraTAG, we propose UltraTAG-S, a robust TAG learning framework designed specifically for sparse scenarios.
(3) \underline{\textit{SOTA Performance.}} 
Our proposed UltraTAG-S achieves SOTA performance and optimal robustness in evaluations among 7 datasets not only in ideal but also in sparse scenarios, exhibiting minimal performance degradation.

\section{Related Work}
\subsection{Graph Learning for Data Sparsity Scenarios}
For graph learning in sparse scenarios, existing research primarily focuses on addressing missing node representations, edge absences, or label deficiencies~\cite{FeaturePropgation,FairAC,gamlp}. Most of them employ vector completion based on graph propagation or attention to handle these issues. However, there is still a lack of targeted research on sparse scenarios of TAGs.

\subsection{Shallow Embedding Methods for TAG Learning}
A common method for TAG learning is shallow embedding, where text attributes are often converted into relatively shallow features like skip-gram~\cite{skip-gram} or BoW~\cite{bow}, which then serve as inputs for traditional graph-based algorithms such as GCN~\cite{kipf2016gcn}. While shallow embeddings are simple and computationally efficient, they fail to capture complex semantics and nuanced relationships with only limited effectiveness.

\subsection{LM/LLM-based Methods for TAG Learning}
With the rise of LMs like BERT~\cite{BERT}, researchers encode textual information in TAGs by fine-tuning LMs on downstream tasks~\cite{GLEM,SimTeG} or aligning LM and GNN via custom loss functions~\cite{G2P2}. The emergence of LLMs like GPT-3~\cite{GPT-3} has further advanced TAG learning, focusing on: (1) text enhancement (e.g., better node descriptions, labels)~\cite{TAPE,GraphBridge,CIKM'24_LLM_as_Teacher,UniGraph}, and (2) superior text encoding for node representations~\cite{ENGINE,GraphAdapter}, collectively boosting overall classification performance.

\begin{figure*}[!htb]
\centering
\includegraphics[width=1.0\linewidth]{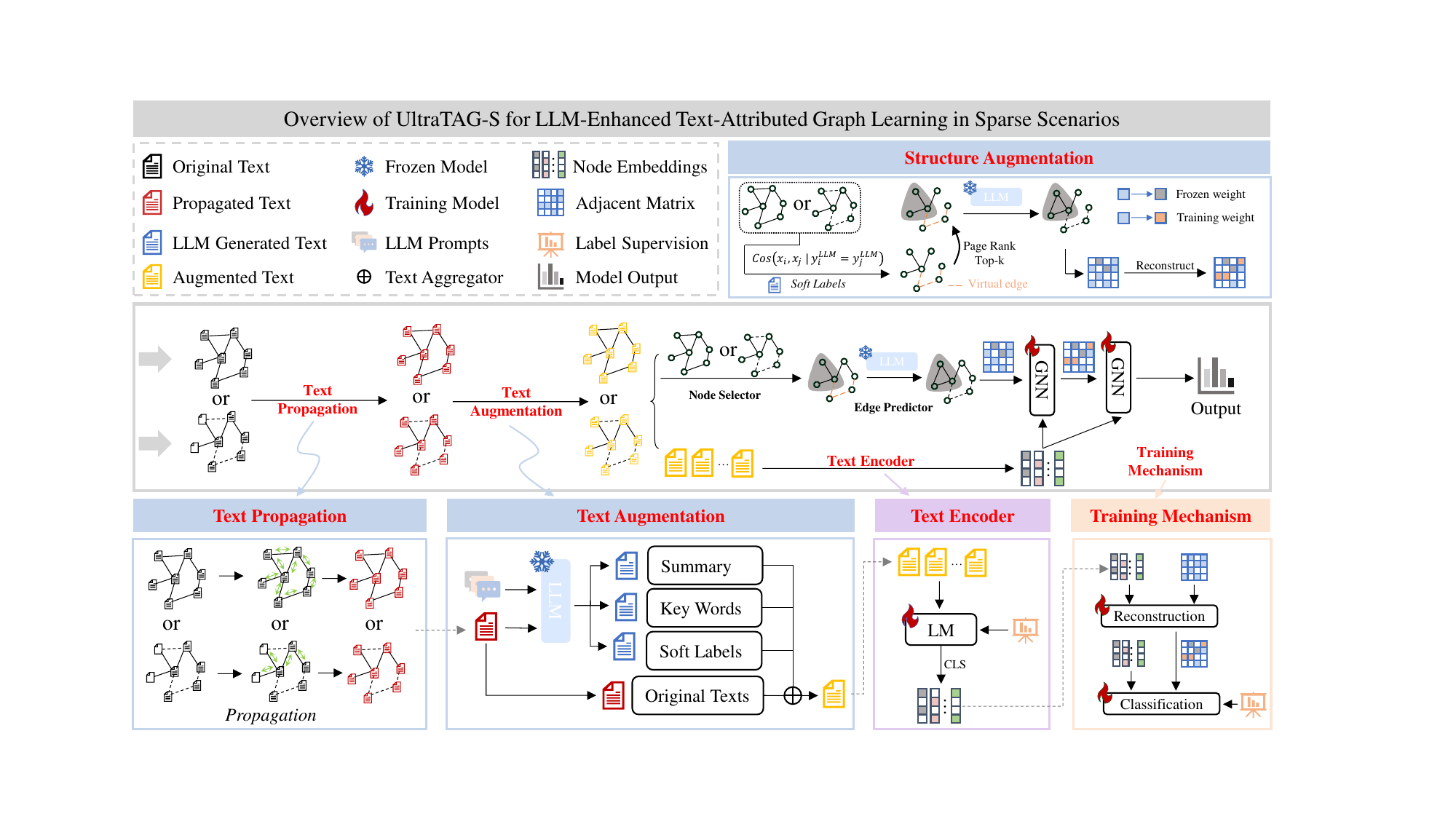}
\caption{{Overview of UltraTAG-S for LLM-Enhanced Text-Attributed Graph Learning in Sparse Scenarios.}}
\label{fig:model2}
\end{figure*}

\section{UltraTAG}
\label{sec:current method}
In this section, we provide details about three organically integrated modules of UltraTAG shown in Figure~\ref{fig:model1}: Data Augmentation, Text Encoder and Training Mechanism.

\subsection{Notations}
Given a TAG $\mathcal{G}=\{ \mathcal{V}, \mathcal{T}, \mathcal{A}, \mathcal{Y}\}$, where $\mathcal{V}$ is the set containing $N$ nodes, $\mathcal{T}$ is the set of texts, for $i \in \mathcal{V}$, $t_i \in \mathcal{T}$ is the text attribute of node $i$. $\mathcal{A} \in \mathbb{R}^{N \times N}$ is the adjacency matrix and $\mathcal{Y}$ is labels. 

This study focuses on the TAG node classification task. The dataset is split into training nodes $\mathcal{V}_{tr}$ with training labels $\mathcal{Y}_{tr}$ and testing nodes $\mathcal{V}_{te}$ with testing labels $\mathcal{Y}_{te}$. A model $f_{\theta^*}$ is trained on $\mathcal{V}_{tr}$ and tested on $\mathcal{V}_{te}$ to generate predictions. The optimization objective is formalized as:
\begin{equation}
f_{\theta^*} = \underset{\theta}{\operatorname*{argmax}}\mathbb{E}_{n \in \mathcal{V}_{tr}}P_\theta(\hat{y}_n=y_n\mid n),
\end{equation}

where $y_n$ is ground-truth and $\hat{y}_n$ is prediction.

\subsection{Data Augmentation}
TAGs rely solely on node texts rather than pre-defined representations, making text preprocessing essential for effective downstream learning. Raw node texts are often noisy, sparse, or semantically limited, and thus cannot directly provide sufficiently rich information for learning expressive representations. Leveraging the powerful generation capabilities of LLMs, we input $\mathcal{T}$ and generate augmented texts $\mathcal{T}^{'}$ using varied prompts $\mathcal{P}$:
\begin{equation}
\mathcal{T}^{'}=\{t^{'}_i\mid t^{'}_i=\text{LLM}(\mathcal{P},t_i,\alpha), \forall t_i\in \mathcal{T}\},
\end{equation}
where $\alpha$ is frozen parameters of LLM.

Inspired by the parameter transmission mechanism in GNNs, we further adopt text propagation to enrich textual representations. After generating $\mathcal{T}^{'}$, they are aggregated with neighboring nodes' texts, allowing contextual information to be shared across the graph, producing the final texts $\mathcal{T}^{*}$:
\begin{equation}
\mathcal{T}^{*} = \{t^{*}_i\mid t^{*}_i= \text{Agg}(t^{'}_i, \{t^{'}_j \mid  j\in \mathcal{N}_i\} )\},
\end{equation}
where $\text{Agg}$ is the text aggregator, which typically include selection, simple concatenation, and other related operations, $\mathcal{N}_i$ is the complete set of immediate neighbor of node $i$.

\subsection{Text Encoder}
The given nodes' texts $\mathcal{T}^{*}_{tr}$ associated with the training nodes must first be encoded into meaningful embeddings to facilitate subsequent model processing, which can generally be efficiently accomplished using the LMs or LLMs.

\textbf{LMs as Encoder.} Text encoding typically employs LMs like BERT~\cite{BERT}. Fine-tuning LMs on downstream tasks enhances their task-specific encoding capability. As for $t_i^{*} \in \mathcal{T}^{*}_{tr}$, this process can be described as:
\begin{equation}
h_i = \text{LM}(t_i^{*},\theta_{\text{LM}}) \in \mathbb{R}^d, \forall t_i^{*}\in \mathcal{T}^{*}_{tr},
\end{equation}
\begin{equation}
\theta^*_{\text{LM}}=\arg\min_{\theta,\theta^*_{\text{LM}}}\sum_{i=1}^N\text{CE}(\text{MLP}(h_i;\theta,\theta_{\text{LM}}),y_i),
\end{equation}
where $h_i$ is the output of the LM, we train the parameters by adding an MLP after the LM with the classification loss.

Meanwhile, various downstream tasks can be used to do it, such as node classification~\cite{SimTeG} or other tailored tasks~\cite{GIANT}.

\textbf{LLMs as Encoder.} Leveraging LLM's language understanding capabilities, their features from different layers capture varying abstraction levels with versatile representations~\cite{ENGINE}. Inspired by it, for each node's text $t_i^{*} \in \mathcal{T}^{*}_{tr} $, we can get $h_i^1, h_i^2, h_i^3, ... , h_i^l$ from different LLM layers:
\begin{equation}
h_i^1, h_i^2, h_i^3, ... , h_i^l = \text{LLM}(t_i^{*},\theta_{\text{LLM}}) \in \mathbb{R}^d,
\end{equation}
where $h_i^j, j\in [1,l]$ denotes the output of LLM layer $j$ of node $i$ and $l$ is the number of LLM layers selected.

LLM's interlayer features with multilevel text representations can also be integrated into GNN modules for message passing, optimizing only GNN parameters:
\begin{equation}
\theta^*=\arg\min_\theta\sum_{i=1}^N \sum_{j=1}^L \text{CE}(\mathcal{M}(h_i^j, \mathcal{A};\theta),y_i),
\end{equation}
where $\mathcal{M}$ denotes the message passing module of GNN, $\mathcal{A}$ is the adjacent matrix, $\text{CE}$ is the Cross-Entropy loss.

\subsection{Training Mechanism}
After obtaining the nodes' textual representations from LM or LLM $\mathcal{H} = \{h_1, h_2, h_3, ..., h_N\}$ and adjacency matrix $\mathcal{A}$, input of them into a GNN will yield the final prediction. In terms of training mechanisms, we can use a simple GNN module, or combine GNN with LM for joint training.

\textbf{Simple GNN.} 
A simple GNN produces final predictions through downstream task training:
\begin{equation}
\theta^*=\arg\min_\theta\sum_{i=1}^N\text{CE}(\text{GNN}(h_i, \mathcal{A};\theta),y_i),
\end{equation}

\textbf{GNN with LM.} 
For the combination of GNN and LM training, the pseudo-labels $\mathcal{Y}_{\text{G}}$ generated by GNN guide LM training, and the pseudo-labels generated by LM $\mathcal{Y}_{\text{L}}$ guide GNN training, and the cycle repeats with the same downstream task:
\vspace{-3pt}
\begin{equation}
\mathcal{Y}_{\text{L}} = \text{GNN}(\mathcal{H}, \mathcal{A}; \theta_{\text{G}}),\mathcal{Y}_{\text{G}} = \text{LM}(\mathcal{T}; \theta_{\text{L}}),
\end{equation}

\section{UltraTAG-S}
\label{sec:our method}
To address the challenge of data sparsity of TAGs in real-word applications, we creatively propose UltraTAG-S as shown in Figure~\ref{fig:model2}, which is composed of three modules: LLM-based Robustness Enhancement, LM-based Resilient Representation Learning and Graph-Enhanced Robust Classifier.
These modules incorporate our unique design for sparse scenarios, such as Text Propagation, Structure Augmentation and Edge Reconfigurator. It should be noted that in this study, we simulate a real-world sparse scenario by deleting nodes' texts and edges in a certain proportion.

\subsection{LLM-based Robustness Enhancement}
Data Augmentation module can be divided into Text Propagation, Text Augmentation and Structure Augmentation.

\underline{\textbf{Text Propagation.}} 
Leveraging the homophily principle in graph theory, we posit that adjacent nodes exhibit textual similarity. Inspired by the message-passing mechanism in GNNs, we propagate textual information from neighboring nodes to reconstruct missing text attributes. Additionally, this propagation method enriches the textual representation of normal nodes, serving as a complementary enhancement.

Specifically, for node $v_i\in \mathcal{V}$, $t_i \in \mathcal{T}$ and its neighbors $\mathcal{N}_i$, propagated texts $\mathcal{T}^{'}$ are obtained by:
\begin{equation}
\mathcal{T}^{'} = \{t^{'}_i\mid t^{'}_i= t_i \oplus \{t_j \mid  j\in \mathcal{N}_i\}, \forall t_i\in \mathcal{T}  \},
\end{equation}
where $\oplus$ denotes concatenation of neighbors' texts.

\underline{\textbf{Text Augmentation.}}
Leveraging the advanced language comprehension of LLMs, we utilize prompt engineering to extract critical textual information and enrich data representations. Traditional preprocessing methods often fail to capture deeper semantic cues or adapt flexibly to diverse graph scenarios. To address this, we developed carefully tailored prompts to guide LLM in generating diverse key texts, including summary, key words, and soft labels. Such enriched textual signals significantly enhance the robustness and informativeness of node representations, providing a stronger foundation for TAG learning tasks.

Specifically, for the propagated text $t_i^{'}\in \mathcal{T}^{'}$, we get the augmented text $\mathcal{T}^{*}$ with different prompts:
\begin{equation}
\mathcal{T}_{\text{Su}} = \{t_i^{''}\mid t_i^{''}=\text{LLM}(\mathcal{P}_\text{Su},t_i^{'},\theta_{\text{LLM}}),
\end{equation}
\begin{equation}
\mathcal{T}_{\text{KW}} = \{t_i^{''}\mid t_i^{''}=\text{LLM}(\mathcal{P}_\text{KW},t_i^{'},\theta_{\text{LLM}}), 
\end{equation}
\begin{equation}
\mathcal{Y}_{\text{SL}} = \{t_i^{''}\mid t_i^{''}=\text{LLM}(\mathcal{P}_\text{SL},t_i^{'},\theta_{\text{LLM}}), 
\end{equation}
\begin{equation}
\mathcal{T}^{*} = \text{AGG}(\mathcal{T}^{'}, \mathcal{T}_{\text{Su}}, \mathcal{T}_{\text{KW}}, \mathcal{Y}_{\text{SL}}),
\end{equation}
where $\text{AGG}$ is text aggregation module of concatenation, $\mathcal{T}_{\text{Su}}$, $\mathcal{T}_{\text{KW}}$, $\mathcal{Y}_{\text{SL}}$ are the Summary, Key Words and Soft Labels generated by LLMs. $\mathcal{P}_\text{Su}$, $\mathcal{P}_\text{KW}$, $\mathcal{P}_\text{SL}$ are the prompts.

\underline{\textbf{Structure Augmentation.}} 
To mitigate edge sparsity, we introduce a Structure Augmentation module composed of Virtual Edge Generator, Node Selector and Edge Reconfigurator. This module leverages LLMs to re-identify edges for selected nodes, thereby optimizing the graph structure. 

\textbf{\textit{a. Virtual Edge Generator.}} To address information sparsity and ensure the graph structure's integrity before node selection, we propose a virtual edge generator. The core idea is to introduce additional edges between semantically related nodes, guided by the soft labels $\mathcal{T}_{\text{SL}}$ generated by LLMs. We first compute node embeddings based on augmented textual representations and then calculate inter-node similarity among those sharing the same soft label:
\begin{equation}
h_i = \text{LM}(t_i^{*},\alpha_{\text{LM}}), h_j = \text{LM}(t_j^{*},\alpha_{\text{LM}}),
\end{equation}
\begin{equation}
\mathcal{S}_{ij} = cos(h_i , h_j), \forall y_i = y_j ~\&~ y_i , y_j \in \mathcal{Y}_{\text{SL}}.
\end{equation}
The similarity scores are then used to enrich the adjacency matrix by adding virtual edges. The adjacency matrix is updated to $\mathcal{A}^{'}$:
\begin{equation}
\mathcal{A}^{'}_{ij} = 
\begin{cases}
1, &\text{if } \mathcal{A}_{ij}=1 | \mathcal{S}_{ij} > \tau_1, \\
0, & \text{else},
\end{cases}
\end{equation}
where $\mathcal{A}$ denotes the adjacency matrix after sparsification and virtual edge addition, while $\tau_1$ is the similarity threshold that determines whether a virtual edge should be established.

\textbf{\textit{b. Node Selector.}} Considering the impracticality of re-evaluating all edges, we design a node selector to identify important nodes $\mathcal{V}_c$. The intuition is that a small subset of important nodes captures the majority of structural and semantic information. We calculate the PageRank score for each node to quantify importance:
\begin{equation}
\text{Score}(v_i) = \text{PageRank}(v_i,\mathcal{A}^{'}),
\end{equation}
\begin{equation}
\mathcal{V}_c = \{v_i \ | \ \text{Score}(v_i) > \text{Score}(v_k) \},
\end{equation}
where $v_k$ is the node with k-th largest node importance score calculated by PageRank with original edges and virtual edges.

\textbf{\textit{c. Edge Reconfigurator.}} For each edge in the induced complete subgraph of $\mathcal{V}_c$, we design an edge reconfiguration process powered by LLMs. The motivation is that structural connections among important nodes should be re-examined with semantic awareness. The LLM evaluates whether an edge should exist and outputs a confidence score $\mathcal{C}{ij}$ for edge $e_{ij}$:
\begin{equation}
\mathcal{C}_{ij} = \text{LLM}(\mathcal{P}_\text{edge},t_i^{*},t_j^{*}), \forall v_i, v_j\in \mathcal{V}_c,
\label{eq:edge reconfiguration}
\end{equation}
The updated adjacency matrix $\mathcal{A}^{*}$ is expressed as:
\vspace{-3pt}
\begin{equation}
\mathcal{A}^{*}_{ij} = 
\begin{cases}
\mathcal{A}_{ij}, & \text{if } v_i \notin \mathcal{V}_c \ | \ v_j \notin \mathcal{V}_c; \\
1, & \text{if }  \mathcal{C}_{ij} > \tau_2;  \quad 0, \ \text{else};
\end{cases}
\end{equation}
where $\tau_2$ denotes the confidence threshold for LLM-guided edge reconfiguration in Equation~\ref{eq:edge reconfiguration}.

\subsection{LM-based Resilient Representation Learning}
After  augmenting the graph $\mathcal{G^*}=\{ \mathcal{V}, \mathcal{T}^*, \mathcal{A}^*, \mathcal{Y}\}$, we fine-tune the language model on downstream tasks of node classification. Specifically, node $i$'s text $t_i^*$ is passed  through the fine-tuned language model $\text{LM}_{\theta}$ to output the feature representation for downstream node classification. The process can be described as following:
\begin{equation}
\hat{y}_i = \text{softmax}(W \cdot \text{LM}(t_i, \theta) + b), \forall \ t_i^* \in \mathcal{T}^*.
\end{equation}

After fine-tuning with the following negative log-likelihood Loss $\mathcal{L}_{ft}$, the node representations $\mathcal{H}$ are calculated by:
\begin{equation}
h_i = \text{LM}(t_i^*, \theta^*),
\end{equation}
\begin{equation}
\mathcal{L}_{ft}=-\frac{1}{N} \sum_{i=1}^N \sum_{k=1}^K y_{i,k}\log\hat{y}_{i,k},
\end{equation}
where $N,K$ are the number of training nodes and classes, $y_{i,k}$ is the ground-truth, $\hat{y}_{i,k}$ is the output. By leveraging this fine-tuning process, the language model provides more resilient textual embeddings that serve as strong foundations for the subsequent graph-based classifier.

\subsection{Graph-Enhanced Robust Classifier}
Considering the inherent sparsity of real-world scenarios, the effectiveness of node propagation in GNN is often compromised due to insufficient edge connectivity. To address this, we adopt a dual-GNN framework for joint training, where the first GNN learns enhanced graph structure representations by capturing higher-order dependencies and the second handles node classification.

Specifically, with the nodes' representations$\mathcal{H} \in \mathbb{R}^{N \times d}$ and the structure representation $\mathcal{A}^{*} \in \mathbb{R}^{N \times N}$, we first compute the similarity matrix of node vector representations:
\begin{equation}
\mathbf{S} = \text{Norm}(\mathcal{H}^{(1)} \cdot \mathcal{H}^{(1)^\top}), \mathcal{H}^{(1)} = \text{GNN}_1(\mathcal{H}, \mathcal{A}^{*}).
\end{equation}

Then, we update the original adjacency matrix $\mathcal{A}^{*}$ with preserving the judgment of the LLM without alteration:
\begin{equation}
\mathcal{\tilde{A}}^{*}_{ij} = 
\begin{cases}
\mathcal{A}^{*}_{ij} + \mathbf{S}_{ij}, & \text{if } v_i \notin \mathcal{V}_c \ | \ v_j \notin \mathcal{V}_c , \\
\mathcal{A}^{*}_{ij}, & \text{else}.
\end{cases}
\end{equation}

We use the updated matrix as the input of $\text{GNN}_2(\cdot)$ and jointly optimize $\text{GNN}_1(\cdot)$ and $\text{GNN}_2(\cdot)$ using cross entropy loss $\mathcal{L}_{GNN}$:
\begin{equation}
\mathcal{H}^{(2)} = \text{GNN}_2(\mathcal{H}, \mathcal{\tilde{A}}^{*}), 
\end{equation}
\begin{equation}
\mathcal{L}_{GNN} = -\frac{1}{N} \sum_{i=1}^N \sum_{k=1}^K y_{i,k}\log\hat{y}_{i,k},
\end{equation}
where $N$ is nodes' number, and $K$ is classes' one. Importantly, during the parameter update process of the graph neural network, we consistently ensure that the node weights of the complete graph formed by important nodes remain unchanged. In other words, we preserve the semantic judgments of the LLM without alteration, thereby maintaining a balance between textual semantics and structural refinement.

\begin{table*}[t]
\centering
\footnotesize
\renewcommand{\arraystretch}{1.1}
\caption{The Comparison of Different LLM-enhanced TAG Learning Methods under Architecture Design and Robustness Dimensions. The top four methods use LMs, while the bottom five use LLMs. 'XMC' is e\underline{X}tream \underline{M}ulti-label \underline{C}lassification, 'Iteration' is iterative training with LM and GNN using pseudo labels, 'Joint' means joint training with multiple GNNs.}
\begin{tabular}{l|>{\centering\arraybackslash}p{2cm}>{\centering\arraybackslash}p{2cm}>{\centering\arraybackslash}p{2.2cm}>{\centering\arraybackslash}p{2.2cm}|>{\centering\arraybackslash}p{1.2cm}>{\centering\arraybackslash}p{1.2cm}>{\centering\arraybackslash}p{1.2cm}>{\centering\arraybackslash}p{1.2cm}}
\toprule
\multirow{2}{*}{Method} & \multicolumn{4}{c|}{Architecture Design} & \multicolumn{4}{c}{Robustness Dimensions} \\ 
\cline{2-9}
& Data Augmentation & Text Encoder & Encoder Supervision & Training Mechanism & Input & Node & Edge & Training \\ 
\hline
GLEM          
& \ding{55} & DeBERTa & Node Classification & Iteration & \ding{55} & \ding{55} & \ding{55} & \ding{55} \\
GIANT        
& \ding{55} & BERT & XMC & Only GNN & \ding{55} & \ding{55} & \ding{55} & \ding{55} \\
G2P2         
& \ding{55} & RoBERTa & Node Classification & Combined Loss & \textcolor{red!50}{\ding{51}} & \textcolor{red!50}{\ding{51}} & \ding{55} & \ding{55} \\
SimTeG      
& \ding{55} & e5-large/RoBERTa & Node Classification & Only GNN & \ding{55} & \ding{55} & \ding{55} & \ding{55} \\
\hline
TAPE         
& \textcolor{red!50}{\ding{51}} & DeBERTa & Node Classification & Only GNN & \textcolor{red!50}{\ding{51}} & \ding{55} & \ding{55} & \ding{55} \\
ENGINE        
& \ding{55} & LLaMA2-7B & / & Joint & \ding{55} & \ding{55} & \ding{55} & \textcolor{red!50}{\ding{51}} \\
LLMGNN      
& \textcolor{red!50}{\ding{51}} & BoW & / & Only GNN & \textcolor{red!50}{\ding{51}} & \ding{55} & \textcolor{red!50}{\ding{51}} & \ding{55} \\
GraphAdapter
& \ding{55} & LLaMA2-13B & Token Prediction & Only GNN & \textcolor{red!50}{\ding{51}} & \ding{55} & \ding{55} & \textcolor{red!50}{\ding{51}} \\
UltraTAG-S
& \textcolor{red!50}{\ding{51}} & BERT & Node Classification & Joint & \textcolor{red!50}{\textcolor{red!50}{\ding{51}}} & \textcolor{red!50}{\textcolor{red!50}{\ding{51}}} & \textcolor{red!50}{\textcolor{red!50}{\ding{51}}} & \textcolor{red!50}{\textcolor{red!50}{\ding{51}}} \\
\bottomrule
\end{tabular}
\label{tab:Comparison of LLM-enhanced TAG Learning Methods}
\end{table*}

\begin{table*}[t]
\centering
\scriptsize
\caption{Experimental results of node classification under different sparsity ratios.
Best is in \colorbox{mycyan!40}{\textbf{bold}} and second in \colorbox{myblue!20}{\underline{underlined}}.}
\setlength{\tabcolsep}{1.4pt}
\begin{tabular}{
w{l}{1cm}|
ccc ccc ccc ccc ccc ccc ccc
}
\toprule
\textbf{Method}
& \multicolumn{3}{c}{Cora}
& \multicolumn{3}{c}{CiteSeer}
& \multicolumn{3}{c}{PubMed}
& \multicolumn{3}{c}{WikiCS}
& \multicolumn{3}{c}{Instagram}
& \multicolumn{3}{c}{Reddit}
& \multicolumn{3}{c}{Elo-Photo} \\

\cmidrule(lr){2-4}\cmidrule(lr){5-7}\cmidrule(lr){8-10}
\cmidrule(lr){11-13}\cmidrule(lr){14-16}\cmidrule(lr){17-19}\cmidrule(lr){20-22}

\textbf{Sparsity}
& 0\% & 50\% & 80\%
& 0\% & 50\% & 80\%
& 0\% & 50\% & 80\%
& 0\% & 50\% & 80\%
& 0\% & 50\% & 80\%
& 0\% & 50\% & 80\%
& 0\% & 50\% & 80\% \\

\midrule

MLP &
54.94 & 36.35 & 30.41 &
61.91 & 40.38 & 27.74 &
52.13 & 49.14 & 44.25 &
62.46 & 34.29 & 26.64 &
51.85 & 62.65 & 62.54 &
52.91 & 52.09 & 50.93 &
47.45 & 46.57 & 46.21 \\

GCN &
74.91 & 64.61 & 41.96 &
69.00 & 48.71 & 30.53 &
72.54 & 63.71 & 49.68 &
73.27 & 64.12 & 54.24 &
63.66 & 58.96 & 63.20 &
55.00 & 55.48 & 54.38 &
69.49 & 62.61 & 51.27 \\

GAT &
71.70 & 60.63 & 38.86 &
70.31 & 51.50 & 30.50 &
75.86 & 66.82 & 52.03 &
68.72 & 64.60 & 52.85 &
64.80 & 64.51 & 63.17 &
60.36 & 56.56 & 56.41 &
64.22 & 59.32 & 50.35 \\

GCNII &
77.23 & 62.36 & 36.79 &
71.91 & 49.66 & 31.07 &
73.28 & 67.01 & 51.33 &
70.12 & 63.62 & 50.83 &
65.07 & 64.08 & 62.27 &
62.78 & 59.27 & 57.62 &
60.60 & 53.77 & 45.53 \\

GraphSAGE &
81.70 & 54.76 & 37.60 &
66.68 & 47.68 & 31.69 &
68.41 & 64.74 & 50.59 &
75.16 & 64.56 & 53.03 &
59.65 & 62.09 & 61.70 &
53.59 & 60.14 & \colorbox{myblue!20}{\underline{58.72}} &
70.48 & 62.61 & 52.17 \\

\midrule
BERT &
79.70 & 55.58 & 37.59 &
76.88 & 48.08 & 31.50 &
90.95 & 66.28 & 49.95 &
71.70 & 45.21 & 28.58 &
63.50 & 63.50 & 62.50 &
58.78 & 54.35 & 51.40 &
70.01 & 57.41 & 49.59 \\

DeBERTa &
73.39 & 34.41 & 29.98 &
75.16 & 43.81 & 30.80 &
90.81 & 65.75 & 42.34 &
68.18 & 32.91 & 21.83 &
62.40 & 63.73 & 63.59 &
59.92 & 50.90 & 50.24 &
70.18 & 57.39 & 47.96 \\

RoBERTa &
80.35 & 53.09 & 28.23 &
77.04 & 44.32 & 23.32 &
91.13 & 66.28 & 47.24 &
72.12 & 42.73 & 20.36 &
64.67 & 63.07 & 63.68 &
59.23 & 50.97 & 50.32 &
70.25 & 57.25 & 49.52 \\

\midrule

GLEM &
87.07 & 64.84 & \colorbox{myblue!20}{\underline{49.01}} &
76.30 & 53.43 & \colorbox{myblue!20}{\underline{36.64}} &
89.56 & 70.51 & 51.48 &
74.83 & 67.07 & 52.41 &
65.90 & 62.43 & 61.54 &
60.88 & 53.85 & 50.82 &
77.74 & 65.25 & 56.25 \\

SimTeG &
88.75 & 72.06 & 45.78 &
77.37 & \colorbox{myblue!20}{\underline{58.11}} & 30.40 &
88.31 & 76.00 & \colorbox{myblue!20}{\underline{54.95}} &
76.32 & 65.34 & 50.35 &
64.29 & 61.55 & 60.61 &
61.60 & 59.84 & 58.08 &
79.82 & 67.76 & 55.73 \\

TAPE &
\colorbox{myblue!20}{\underline{89.07}} & \colorbox{myblue!20}{\underline{78.73}} & 47.08 &
77.02 & 54.31 & 29.77 &
90.38 & \colorbox{myblue!20}{\underline{77.18}} & 54.87 &
80.17 & \colorbox{myblue!20}{\underline{73.62}} & \colorbox{myblue!20}{\underline{59.83}} &
65.44 & 61.40 & 61.25 &
\colorbox{myblue!20}{\underline{63.01}} & 60.18 & 58.10 &
82.26 & \colorbox{myblue!20}{\underline{76.21}} & \colorbox{myblue!20}{\underline{59.76}} \\

ENGINE &
86.79 & 70.85 & 42.32 &
\colorbox{myblue!20}{\underline{78.03}} & 56.30 & 35.70 &
\colorbox{myblue!20}{\underline{91.43}} & 75.42 & 54.74 &
\colorbox{myblue!20}{\underline{81.38}} & 71.72 & 49.42 &
\colorbox{myblue!20}{\underline{66.27}} & \colorbox{myblue!20}{\underline{64.74}} & \colorbox{myblue!20}{\underline{63.88}} &
62.57 & \colorbox{myblue!20}{\underline{60.18}} & 57.54 &
\colorbox{myblue!20}{\underline{83.06}} & 73.40 & 57.96 \\
\midrule
UltraTAG-S &
\colorbox{mycyan!40}{\textbf{90.96}} & \colorbox{mycyan!40}{\textbf{83.95}} & \colorbox{mycyan!40}{\textbf{57.57}} &
\colorbox{mycyan!40}{\textbf{78.68}} & \colorbox{mycyan!40}{\textbf{67.08}} & \colorbox{mycyan!40}{\textbf{40.08}} &
\colorbox{mycyan!40}{\textbf{92.41}} & \colorbox{mycyan!40}{\textbf{80.91}} & \colorbox{mycyan!40}{\textbf{61.05}} &
\colorbox{mycyan!40}{\textbf{83.05}} & \colorbox{mycyan!40}{\textbf{77.45}} & \colorbox{mycyan!40}{\textbf{65.60}} &
\colorbox{mycyan!40}{\textbf{66.69}} & \colorbox{mycyan!40}{\textbf{65.61}} & \colorbox{mycyan!40}{\textbf{64.78}} &
\colorbox{mycyan!40}{\textbf{63.78}} & \colorbox{mycyan!40}{\textbf{60.34}} & \colorbox{mycyan!40}{\textbf{59.85}} &
\colorbox{mycyan!40}{\textbf{84.70}} & \colorbox{mycyan!40}{\textbf{79.21}} & \colorbox{mycyan!40}{\textbf{68.79}} \\

\bottomrule
\end{tabular}
\label{tab:experiment_results}
\end{table*}

\section{Experiments}
In this section, we analyze the effectiveness of UltraTAG-S through experimental evaluation. To comprehensively assess the performance of our approach, we address the following research questions:
\textbf{\textit{Q1: }}What are the key differences between existing LLM-enhanced TAG learning methods when applied under the UltraTAG framework?
\textbf{\textit{Q2: }}What is the overall performance of UltraTAG-S as a general and robust TAGs learning paradigm in both ideal and sparse data scenarios?
\textbf{\textit{Q3: }}What critical factors and design choices contribute to the superior performance and robustness of UltraTAG-S?
\textbf{\textit{Q4: }}What are the computational complexity and hyperparameter settings of training UltraTAG-S in practice?


\begin{table}[t]
\centering
\footnotesize
\caption{Statistics of the TAG datasets. The datasets are partitioned in Train-Val-Test-Out mode, 'Out' is data that not involved in any of training, validation, or test sets.}
{\begin{tabular}{w{l}{1.5cm}|w{c}{1.2cm}w{c}{1.2cm}w{c}{1.2cm}w{c}{1cm}w{c}{2.5cm}c}
\toprule
Dataset          & \#Nodes & \#Edges & \#Classes &\#Split Ratio(\%) \\  
\midrule
Cora~\cite{collective}             & 2,708     & 5,278      & 7     & 60-20-20-0       \\
CiteSeer~\cite{citeseer}         & 3,186     & 4,277      & 6     & 60-20-20-0       \\
PubMed~\cite{collective}           & 19,717    & 44,324     & 3     & 60-20-20-0       \\
WikiCS~\cite{mernyei2020wiki}           & 11,701    & 215,863    & 10    & 5-15-50-30       \\
Instagram~\cite{GraphAdapter}        & 11,339    & 144,010    & 2     & 10-10-80-0      \\
Reddit~\cite{GraphAdapter}           & 33,434    & 198,448    & 2     & 10-10-80-0       \\
Elo-Photo~\cite{yan2023comprehensive}        & 48,362    & 873,793    & 12    & 40-15-45-0       \\
\bottomrule
\end{tabular}}
\label{tab:Statistics of the TAG datasets.}
\end{table}

\subsection{Paradigm Comparision}
In this section, we compare current LLM-enhanced TAG learning methods under UltraTAG from four aspects: Data Augmentation, Text Encoder, Encoder Supervision, and Training Mechanism, as shown in Table~\ref{tab:Comparison of LLM-enhanced TAG Learning Methods}. This comparison clarifies the positioning of each method, reveals their limitations and complementarities, and motivates the design of UltraTAG-S.

As for LM-based methods GLEM~\cite{GLEM}, GIANT~\cite{GIANT}, G2P2~\cite{G2P2} and SimTeG~\cite{SimTeG}, they respectively employed different LMs and designed diverse fine-tuning tasks with varying supervision strategies. As for LLM-based methods, only TAPE~\cite{TAPE}, LLMGNN~\cite{LLMGNN} and UltraTAG-S perform explicit data augmentation for enhanced representations. Meanwhile, anyone can flexibly optimize one small module among the four modules of UltraTAG to form a new baseline method.

We also compare the robustness of different LLM-enhanced TAG learning methods in sparse scenarios, examining whether they consider robustness of input data, missing texts, missing edges, and unstable training dynamics. Only UltraTAG-S consistently addresses robustness across all dimensions, highlighting its superiority in real-world environments.

\subsection{Performance and Robustness Analysis}

\begin{figure*}[!htb]
\centering
\includegraphics[width=0.9\linewidth]{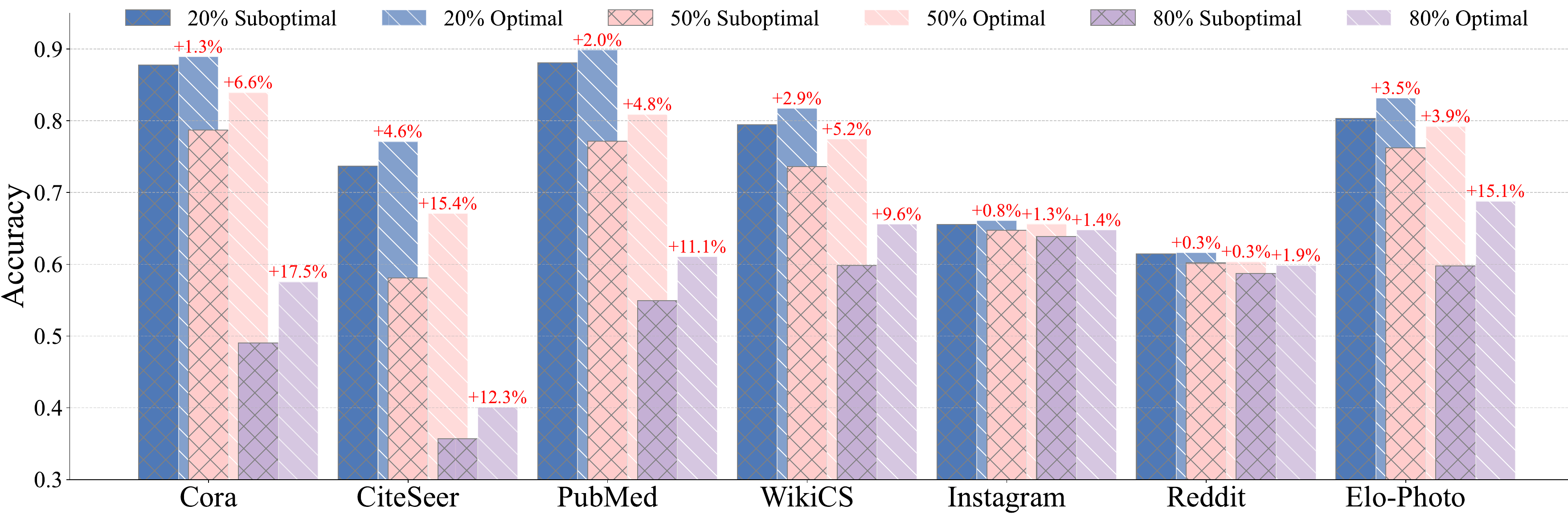}
\caption{Robustness Comparison among All Datasets in Sparse Ratio of 20\%, 50\% and 80\%.}
\label{fig:Robustness Comparison among All Datasets in Sparse Ratio of 20, 50 and 80.}
\end{figure*}

We conduct a comprehensive evaluation of UltraTAG-S by comparing it with GNN-only, LM-only, and LLM–GNN hybrid methods among different TAG datasets, as reported in Table~\ref{tab:experiment_results}. The statistics of the TAG datasets are in Table~\ref{tab:Statistics of the TAG datasets.}.
Since GNN-only methods cannot directly accept raw texts, to ensure a fair comparison we first encode node texts with a unified BERT~\cite{BERT} and feed the resulting representations as inputs to these models.
To simulate sparse scenarios, we randomly remove node texts and edges at ratios of 20\%, 50\%, and 80\% without extra denoising or constraints. 
As can be seen from Table~\ref{tab:experiment_results}, UltraTAG-S achieves the best performance on all datasets among different sparsity ratios, with the maximum gain reaching 2.21\% over the strongest baseline in normal scenario and 17.5\% in sparse scenarios. LM-based classifiers consistently outperform GNN-only methods on TAG datasets, and LLM–GNN integration further boosts performance by combining semantic and structural reasoning. UltraTAG-S extends this integration with LLM-guided augmentation, structure refinement, and dual-GNN training, yielding consistent accuracy and stability improvements across domains, scales, and noise settings.

Figure~\ref{fig:Robustness Comparison among All Datasets in Sparse Ratio of 20, 50 and 80.} and Figure~\ref{fig:Robustness Comparison and Homophily Analysis}(Left) further demonstrate the robustness of UltraTAG-S across different sparsity levels. UltraTAG-S consistently outperforms suboptimal baselines across all datasets and sparsity settings, and exhibits the smallest accuracy degradation under increasing sparsity. This robustness arises from LLM-based text augmentation and propagation, virtual-edge construction to restore connectivity, and LLM-guided edge reconfiguration that preserves critical semantics. Even at 80\% sparsity, UltraTAG-S achieves state-of-the-art performance with gains up to 17.5\%, and its relative advantage grows as sparsity increases, highlighting its effectiveness in addressing key failure modes of TAG learning under incomplete texts and disrupted graph structures.

\subsection{Ablation Study}
\begin{figure*}[!htb]
\centering
\includegraphics[width=\linewidth]{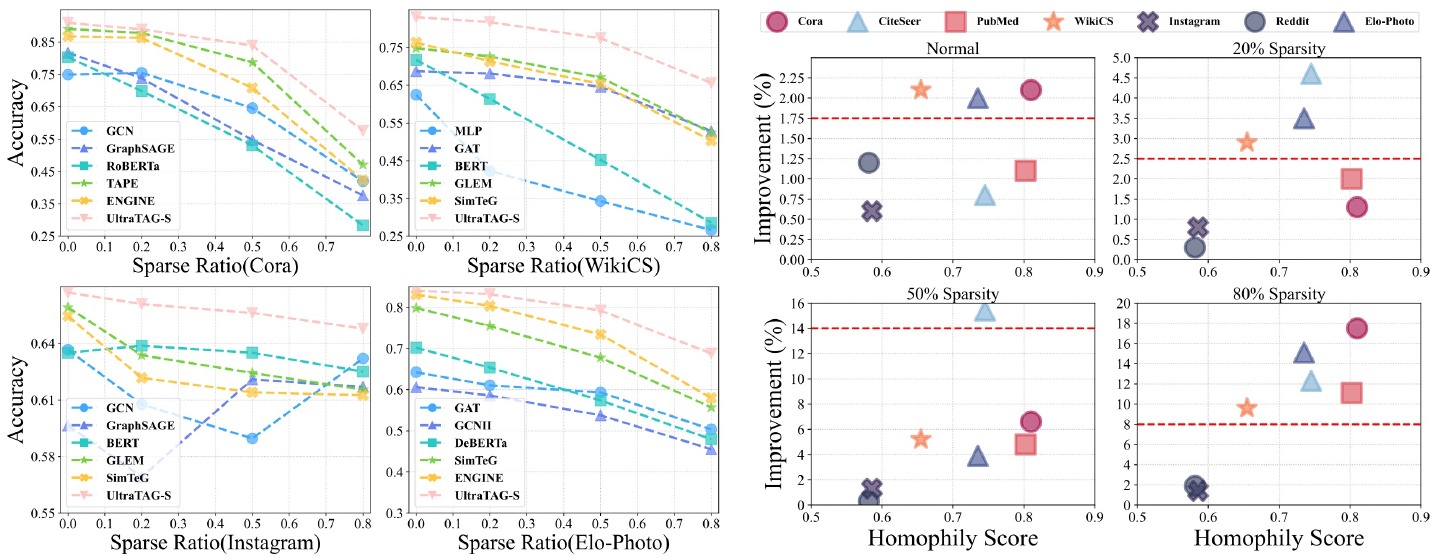}
\caption{Left: Robustness Comparison in Sparse Scenarios among Four Datasets. Right: Homophily Analysis of the Datasets. The horizontal coordinate is the homophily score of the datasets, and the vertical coordinate is the improvement of UltraTAG-S compared to the suboptimal method.}
\label{fig:Robustness Comparison and Homophily Analysis}
\end{figure*}

We conduct an ablation study on Cora~\cite{collective}, CiteSeer~\cite{citeseer}, and PubMed~\cite{collective} to evaluate the effectiveness and robustness of UltraTAG-S under sparse scenarios, with results reported in Table~\ref{tab:Ablation Study.}. Each module contributes substantially to overall performance and robustness. Text Augmentation improves generalization by enriching textual diversity, yielding gains of up to 18.46\% on Cora, 16.89\% on CiteSeer, and 55.45\% on PubMed. Structure Augmentation enhances robustness through graph topology optimization, bringing additional improvements of 2.44\%, 3.07\%, and 5.90\%, respectively. Structure Learning delivers the most significant gains of 51.10\% on Cora, 32.49\% on CiteSeer, and 40.09\% on PubMed—demonstrating its critical effectiveness in capturing complex graph dependencies. These results collectively highlight the complementary nature of all modules and the paramount importance of structure learning for achieving robust TAG performance under sparse conditions. Notably, the contribution of each module becomes increasingly pronounced as sparsity increases, further confirming that all components are indispensable for maintaining strong performance in severely degraded environments.

\aptLtoX[graphic=no,type=html]{\begin{table}[htpb]
\caption{Detailed Performance Comparison of Ablation Study. 'w/o TA' means without Text Augmentation module, 'w/o SA' means without Structure Augmentation module, 'w/o SL' means without Structure Learning module.}
\label{tab:Ablation Study.}
\centering
\footnotesize
\scalebox{0.9}{
\begin{tabular}{w{l}{2cm}|w{c}{1.7cm}w{c}{1.7cm}w{c}{1.7cm}c}
\toprule
Method          & Cora & CiteSeer & PubMed \\
\midrule
\cellcolor[HTML]{e5e5e5}{Sparse Ratio}  & \cellcolor[HTML]{e5e5e5}{0\%} & \cellcolor[HTML]{e5e5e5}{0\%} & \cellcolor[HTML]{e5e5e5}{0\%} \\
w/o TA	&89.59±1.39	&77.37±0.96	&91.29±0.90\\
w/o SA	&90.59±0.86	&78.37±1.63	&91.99±1.75\\
w/o SL	&88.38±0.85	&77.59±1.44	&86.36±1.23\\
UltraTAG-S	&\textbf{90.96±0.45}	&\textbf{78.68±0.21}	&\textbf{92.41±0.30}\\
\midrule
\cellcolor[HTML]{e5e5e5}{Sparse Ratio}  & \cellcolor[HTML]{e5e5e5}{20\%} & \cellcolor[HTML]{e5e5e5}{20\%} & \cellcolor[HTML]{e5e5e5}{20\%} \\
w/o TA	&86.52±0.21	&75.12±0.45	&88.38±1.50\\
w/o SA	&87.93±1.72	&77.01±0.31	&89.18±0.58\\
w/o SL	&86.72±1.33	&68.34±0.90	&85.42±1.34\\
UltraTAG-S	&\textbf{88.93±0.74}	&\textbf{77.12±0.28}	&\textbf{89.88±0.21}\\
\midrule
\cellcolor[HTML]{e5e5e5}{Sparse Ratio}  & \cellcolor[HTML]{e5e5e5}{50\%} & \cellcolor[HTML]{e5e5e5}{50\%} & \cellcolor[HTML]{e5e5e5}{50\%} \\
w/o TA	&78.41±0.32	&59.94±1.07	&52.05±0.45\\
w/o SA	&81.95±1.30	&65.08±1.06	&78.76±0.22\\
w/o SL	&74.54±0.54	&50.63±1.42	&69.17±0.71\\
UltraTAG-S	&\textbf{83.95±0.80}	&\textbf{67.08±0.28}	&\textbf{80.91±0.29}\\
\midrule
\cellcolor[HTML]{e5e5e5}{Sparse Ratio}  & \cellcolor[HTML]{e5e5e5}{80\%} & \cellcolor[HTML]{e5e5e5}{80\%} & \cellcolor[HTML]{e5e5e5}{80\%} \\
w/o TA	&48.60±1.83	&34.29±1.42	&44.93±1.22\\
w/o SA	&56.83±1.45	&39.50±0.54	&57.65±1.21\\
w/o SL	&38.01±1.08	&35.71±0.04	&43.58±0.45\\
UltraTAG-S	&\textbf{57.57±1.38}	&\textbf{40.08±0.45}	&\textbf{61.05±0.49}\\
\bottomrule
\end{tabular}
}
\end{table}}{\begin{table}[htpb]
\caption{Detailed Performance Comparison of Ablation Study. 'w/o TA' means without Text Augmentation module, 'w/o SA' means without Structure Augmentation module, 'w/o SL' means without Structure Learning module.}
\label{tab:Ablation Study.}
\centering
\footnotesize
\scalebox{0.9}{
\begin{tabular}{w{l}{2cm}|w{c}{1.7cm}w{c}{1.7cm}w{c}{1.7cm}c}
\toprule
Method          & Cora & CiteSeer & PubMed \\
\midrule
\rowcolor{Gray}
Sparse Ratio  & 0\% & 0\% & 0\% \\
w/o TA	&89.59±1.39	&77.37±0.96	&91.29±0.90\\
w/o SA	&90.59±0.86	&78.37±1.63	&91.99±1.75\\
w/o SL	&88.38±0.85	&77.59±1.44	&86.36±1.23\\
UltraTAG-S	&\textbf{90.96±0.45}	&\textbf{78.68±0.21}	&\textbf{92.41±0.30}\\
\midrule
\rowcolor{Gray}
Sparse Ratio  & 20\% & 20\% & 20\% \\
w/o TA	&86.52±0.21	&75.12±0.45	&88.38±1.50\\
w/o SA	&87.93±1.72	&77.01±0.31	&89.18±0.58\\
w/o SL	&86.72±1.33	&68.34±0.90	&85.42±1.34\\
UltraTAG-S	&\textbf{88.93±0.74}	&\textbf{77.12±0.28}	&\textbf{89.88±0.21}\\
\midrule
\rowcolor{Gray}
Sparse Ratio  & 50\% & 50\% & 50\% \\
w/o TA	&78.41±0.32	&59.94±1.07	&52.05±0.45\\
w/o SA	&81.95±1.30	&65.08±1.06	&78.76±0.22\\
w/o SL	&74.54±0.54	&50.63±1.42	&69.17±0.71\\
UltraTAG-S	&\textbf{83.95±0.80}	&\textbf{67.08±0.28}	&\textbf{80.91±0.29}\\
\midrule
\rowcolor{Gray}
Sparse Ratio  & 80\% & 80\% & 80\% \\
w/o TA	&48.60±1.83	&34.29±1.42	&44.93±1.22\\
w/o SA	&56.83±1.45	&39.50±0.54	&57.65±1.21\\
w/o SL	&38.01±1.08	&35.71±0.04	&43.58±0.45\\
UltraTAG-S	&\textbf{57.57±1.38}	&\textbf{40.08±0.45}	&\textbf{61.05±0.49}\\
\bottomrule
\end{tabular}
}
\end{table}}

Considering that the homogeneity and heterogeneity of the TAG dataset may impact the text and structure augmentation modules, we examine this effect in Figure~\ref{fig:Robustness Comparison and Homophily Analysis}(Right). We compute homophily scores for all datasets and measure performance gains over the suboptimal method under both normal and sparse settings. UltraTAG-S consistently improves performance across all datasets. Notably, higher homophily leads to larger gains, as semantically similar neighbors enhance augmentation and structure refinement. The best improvement occurs around a homophily score of 0.7, indicating that UltraTAG-S effectively exploits moderate structural consistency while remaining robust in less homogeneous graphs.

\begin{table}[!h]
\centering
\caption{The comparison of different methods in downstream GNN training time per epoch.}
{\begin{tabular}{p{2.2cm}|>{\centering\arraybackslash}p{1.5cm}>{\centering\arraybackslash}p{1.5cm}>{\centering\arraybackslash}p{1.5cm}}
\toprule
& Cora   & PubMed & WikiCS \\
\midrule
SimTeG     & 0.142s & 1.564s & 2.109s \\
GLEM       & 0.131s & 1.034s & 1.645s \\
ENGINE     & 0.290s & 2.165s & 2.730s \\
\midrule
UltraTAG-S & \colorbox{mycyan!40}{\textbf{0.015s}} & \colorbox{mycyan!40}{\textbf{0.170s}} & \colorbox{mycyan!40}{\textbf{0.845s}} \\
\bottomrule
\end{tabular}}
\label{tab:Efficiency Analysis.}
\end{table}

\subsection{Complexity Analysis and Hyperparameter Settings}
The computational complexity of our proposed method is primarily determined by two GNN operations. The first GNN calculates the similarity matrix $\mathbf{S} \in \mathbb{R}^{N \times N}$ and updates the adjacency matrix $\mathcal{A}^{*}_{ij}$. This step involves pairwise computations between nodes, leading to a complexity of $O(N^2)$. The second GNN performs node classification using the updated adjacency matrix $\mathcal{\tilde{A}}^{*}_{ij}$ and node features $\mathcal{H}$. With $m$ layers and $\mathcal{E} = O(N^2)$ edges in the graph, the complexity for this operation scales as $O(m \cdot N^2)$.
Therefore, the total computational cost per epoch is dominated by these two steps, resulting in an overall complexity of $O(m \cdot N^2)$.

Experimentally, compared to other TAG learning methods, UltraTAG-S demonstrates a significant advantage in training time. As shown in Table~\ref{tab:Efficiency Analysis.}, under non-sparse settings, it achieves up to a 19× speedup per training epoch over the suboptimal method. This efficiency gain is particularly important for large-scale TAG datasets, where traditional LLM–GNN hybrids often suffer from excessive overhead. UltraTAG-S benefits from its streamlined dual-GNN framework and lightweight LLM integration, which avoids redundant computations while providing rich semantic and structural enhancements. Consequently, our method scales more gracefully with increasing dataset size and sparsity levels, offering both practical efficiency and theoretical robustness. This makes UltraTAG-S well-suited for real-world deployment where computational resources may be limited while the demand for robust performance under data imperfections remains high. Moreover, the preprocessing stage involving LLM-based augmentation is performed only once offline, further reducing the overall computational burden during training.

For detailed parameter settings, we employ five random seeds {42, 43, 44, 45, 46} to ensure the stability and reproducibility of our experimental results. The $\text{GNN}_1(\cdot)$ and $\text{GNN}_2(\cdot)$ use the Adam optimizer for joint optimizing, with a learning rate of 1e-2, weight decay of 5e-4, dropout of 0.5, and a total of 100 training epochs, which strikes a good balance between convergence speed and generalization ability. Each GNN consists of 2 layers, with a similarity calculation threshold of 0.8 to control the reliability of edge updates. The number of important nodes selected by PageRank is set to 10\% of total training nodes, while the acceptance threshold for LLM-based edge reconfiguration is 0.5, ensuring that only semantically meaningful structural modifications are retained. For fine-tuning the LM, the learning rate is fixed at 5e-5, with 3 epochs, a batch size of 8, and a dropout of 0.3, which are empirically chosen to provide stable training without overfitting. The LLM used is Meta-Llama-3-8B-Instruct~\cite{llama3modelcard}, applied in full compliance with its research license terms. All experiments were conducted on a high-performance system equipped with a single NVIDIA A100 80GB PCIe GPU. All reported results are averaged over five runs with the above seeds, and error bars are included to indicate statistical reliability.

{

}

\section{Conclusion}
In response to the current LLM-enhanced TAG Learning methods, we first propose UltraTAG as a unified and domain-adaptive pipeline learning framework. Simultaneously, to address the challenges faced by existing methods in real-world sparse scenarios, we introduce UltraTAG-S, a TAG learning paradigm specifically tailored for sparse scenarios. UltraTAG-S effectively resolves the issues of text sparsity and edge sparsity through LLM-based text propagation and augmentation strategies, as well as PageRank and LLM-based graph structure learning strategies. 

\begin{acks}
This work was partially supported by the NSFC Grants 625B2021, U24A20255, U2241211,  and 62427808.
\end{acks}

\pagebreak
\bibliographystyle{ACM-Reference-Format}
\bibliography{sample-base}

\end{document}